\documentstyle[colap]{article}

\makeatletter
\def\section{\@startsection {section}{1}{\z@}
  {-2ex plus -.3ex minus  -.1ex}{1ex plus .1ex}{\Large\bf}}
\def\subsection{\@startsection{subsection}{2}{\z@}
  {-1.5ex plus -.4ex minus -.1ex}{1ex plus .1ex}{\large\bf}}
\long\def\@makecaption#1#2{
 \vskip .2ex 
 \setbox\@tempboxa\hbox{#1: #2}
 \ifdim \wd\@tempboxa >\hsize #1: #2\par \else \hbox
to\hsize{\hfil\box\@tempboxa\hfil} 
 \fi}
\makeatother

\newcommand{\pos}{{\sc pos}}
\newcommand{\ignore}[1]{}
\newcommand{\ra}{$\rightarrow$}

\newcommand{\andrei}{\scriptsize\bf \setlength{\parindent}{0mm} }
\newcommand{\longline}{\rule{\textwidth}{0.01in}}

\title{\vspace{-0.5in} 
         Learning Part-of-Speech Guessing Rules from Lexicon: \\ 
    Extension to Non-Concatenative Operations\thanks{ some of the
    research reported here was funded as part of {\sc epsrc} project
    IED4/1/5808 ``Integrated Language Database''.} }

\author{Andrei Mikheev  \\
\\  HCRC Language Technology Group \\ University of Edinburgh \\
2 Buccleuch Place \\ Edinburgh EH8 9LW, Scotland, UK \\ 
{\tt: Andrei.Mikheev@ed.ac.uk}}

\begin{document}
\bibliographystyle{fullname}
\maketitle
\vspace{-0.5in}
\begin{abstract}
  One of the problems in part-of-speech tagging of real-word texts is
  that of unknown to the lexicon words. In \cite{Mikheev:1996}, a
  technique for fully unsupervised statistical acquisition of rules which
  guess possible parts-of-speech for unknown words was proposed.  One of
  the over-simplification assumed by this learning technique was the
  acquisition of morphological rules which obey only simple concatenative
  regularities of the main word with an affix. In this paper we extend
  this technique to the non-concatenative cases of suffixation and assess
  the gain in the performance.
\end{abstract}

\section{Introduction}

Part-of-speech (\pos ) taggers are programs which assign a single \pos
-tag to a word-token, provided that it is known what parts-of-speech this
word can take on in principle. In order to do that taggers are supplied
with a lexicon that lists possible \pos -tags for words which were seen
at the training phase. Naturally, when tagging real-word texts, one can
expect to encounter words which were not seen at the training phase and
hence not included into the lexicon.  This is where word-\pos\ guessers
take their place - they employ the analysis of word features, e.g.  word
leading and trailing characters to figure out its possible \pos\ 
categories.  Currently, most of the taggers are supplied with a
word-guessing component for dealing with unknown words. The most popular
guessing strategy is so-called ``ending guessing'' when a possible set of
\pos -tags for a word is guessed solely on the basis of its trailing
characters. An example of such guesser is the guesser supplied with the
Xerox tagger \cite{Kupiec:1992}.  A similar approach was taken in
\cite{Weischedel:1993} where an unknown word was guessed given the
probabilities for an unknown word to be of a particular \pos , its
capitalisation feature and its ending.  In \cite{Brill:1995} a system of
rules which uses both ending-guessing and more morphologically motivated
rules is described.  Best of these methods were reported to achieve
82--85\% of tagging accuracy on unknown words, e.g.
\cite{Brill:1995,Weischedel:1993}.

In \cite{Mikheev:1996} a cascading word-\pos\ guesser is described.  It
applies first morphological prefix and suffix guessing rules and then
ending-guessing rules. This guesser is reported to achieve higher
guessing accuracy than quoted before which in average was about by 8-9\%
better than that of the Xerox guesser and by 6-7\% better than that of
Brill's guesser, reaching 87-92\% tagging accuracy on unknown words.

There are two kinds of word-guessing rules employed by the cascading
guesser: morphological rules and ending guessing rules.  Morphological
word-guessing rules describe how one word can be guessed given that
another word is known.  In English, as in many other languages,
morphological word formation is realised by affixation: prefixation and
suffixation, so there are two kinds of morphological rules: suffix rules
($A^{s}$) --- rules which are applied to the tail of a word, and prefix
rules ($A^{p}$) --- rules which are applied to the beginning of a word.
For example, the prefix rule:

$A^{p}$ : {\andrei [un (VBD VBN) (JJ)] } 

says that if segmenting the prefix ``un'' from an unknown word results in
a word which is found in the lexicon as a past verb and participle
{\andrei (VBD VBN)}, we conclude that the unknown word is an adjective
(JJ). This rule works, for instance, for words {\andrei [developed \ra
  undeveloped]}.  An example of a suffix rule is:

$A^{s}$ :  {\andrei [ed (NN VB) (JJ VBD VBN)]}

This rule says that if by stripping the suffix ``ed'' from an unknown
word we produce a word with the \pos -class noun/verb {\andrei (NN VB)},
the unknown word is of the class adjective/past-verb/participle {\andrei
  (JJ VBD VBN)}.  This rule works, for instance, for word pairs {\andrei
  [book \ra booked], [water \ra watered]}, etc.

Unlike morphological guessing rules, ending-guessing rules do not require
the main form of an unknown word to be listed in the lexicon.  These
rules guess a \pos -class for a word just on the basis of its ending
characters and without looking up its stem in the lexicon.  For example,
an ending-guessing rule

$A^{e}$:{\andrei  [ing ---  (JJ NN VBG)]}

says that if a word ends with ``ing'' it can be an adjective, a noun or a
gerund. Unlike a morphological rule, this rule does not ask to check
whether the substring preceeding the ``ing''-ending is a word with a
particular \pos -tag.

Not surprisingly, morphological guessing rules are more accurate than
ending-guessing rules but their lexical coverage is more restricted, i.e.
they are able to cover less unknown words.  Since they are more accurate,
in the cascading guesser they were applied before the ending-guessing
rules and improved the precision of the guessings by about 5\%. This,
actually, resulted in about 2\% higher accuracy of tagging on unknown
words.

Although in general the performance of the cascading guesser was detected
to be only 6\% worse than a general-language lexicon lookup, one of the
over-simplifications assumed at the extraction of the morphological rules
was that they obey only simple concatenative regularities:

{\andrei book \ra book+ed; take \ra take+n; play \ra play+ing}.

No attempts were made to model non-concatenative cases which are quite
common in English, as for instance:

{\andrei try \ra tries; reduce\ra reducing; advise\ra advisable}.

So we thought that the incorporation of a set of guessing rules which can
capture morphological word dependencies with letter alterations should
extend the lexical coverage of the morphological rules and hence might
contribute to the overall guessing accuracy.

In the rest of the paper first, we will briefly outline the unsupervised
statistical learning technique proposed in \cite{Mikheev:1996}, then we
propose a modification which will allow for the incorporation of the
learning of non-concatenative morphological rules, and finally, we will
evaluate and assess the contribution of the non-concatenative suffix
morphological rules to the overall tagging accuracy on unknown words
using the cascading guesser.

\section{The Learning Paradigm}

The major topic in the development of word-\pos\ guessers is the strategy
which is to be used for the acquisition of the guessing rules.  
\ignore{
  A rule-based tagger described in \cite{Voutilainen:1995} is equipped
  with a set of guessing rules which has been hand-crafted using
  knowledge of English morphology and intuition.  A more appealing
  approach is an empirical automatic acquisition of such rules using
  available lexical resources.  In \cite{Zhang:1990} a system for the
  automated learning of morphological word-formation rules is described.
  This system divides a string into three regions and from training
  examples infers their correspondence to underlying morphological
  features.  } 
Brill \cite{Brill:1995} outlines a transformation-based
learner which learns guessing rules from a pre-tagged training corpus.  A
statistical-based suffix learner is presented in \cite{Schmid:1994}. From
a pre-tagged training corpus it constructs the suffix tree where every
suffix is associated with its information measure.

The learning technique employed in the induction of the rules of the
cascading guesser \cite{Mikheev:1996} does not require specially prepared
training data and employs fully unsupervised statistical learning from
the lexicon supplied with the tagger and word-frequencies obtained from a
raw corpus. The learning is implemented as a two-staged process with
feedback. First, setting certain parameters a set of guessing rules is
acquired, then it is evaluated and the results of evaluation are used for
re-acquisition of a better tuned rule-set. As it has been already said,
this learning technique proved to be very successful, but did not attempt
at the acquisition of word-guessing rules which do not obey simple
concatenations of a main word with some prefix.  Here we present an
extension to accommodate such cases.

\subsection {Rule Extraction Phase}
\label{sec:morph}

In the initial learning technique \cite{Mikheev:1996} which accounted
only for simple concatenative regularities a guessing rule was seen as a
triple: $A=(S,I,R)$ where

$S$ is the affix itself;

$I$ is the \pos -class of words which should be looked up in the
  lexicon as main forms; 

$R$ is the \pos -class which is assigned to unknown words if the
rule is satisfied.

Here we extend this structure to handle cases of the mutation in the last
$n$ letters of the main word (words of $I$-class), as, for instance, in
the case of {\andrei try \ra tries}, when the letter ``y'' is changed to
``i'' before the suffix.  To accommodate such alterations we included an
additional {\em mutation} element ($M$) into the rule structure.  This
element keeps the segment to be added to the main word.  So the
application of a guessing rule can be described as:

unknown-word - $S$ + $M$ : $I$ \ra $R$ \\
i.e. from an unknown word we strip the affix $S$, add the mutative segment
$M$, lookup the produced string in the lexicon and if it is of  class $I$
we conclude that the unknown word is of  class $R$.
For example: the suffix rule
$A^{s}$:

[ $S$={\andrei ied} $I$={\andrei (NN, VB)} $R$={\andrei (JJ VBD  VBN)} $M$=y]

or in short {\andrei [ied (NN VB) (JJ VBD VBN) y]}\\ says that if there
is an unknown word which ends with ``ied'', we should strip this ending
and append to the remaining part the string ``y''. If then we find this
word in the lexicon as {\andrei (NN VB)} (noun/verb), we conclude that
the guessed word is of category {\andrei (JJ VBD VBN)} (adjective, past
verb or participle). This rule, for example, will work for word pairs
like {\andrei specify - specified} or {\andrei deny - denied}.

Next, we modified the $\bigtriangledown$ operator which was used for the
extraction of morphological guessing rules. We augmented this operator
with the index $n$ which specifies the length of the mutative ending of
the main word. Thus when the index $n$ is 0 the result of the application
of the $\bigtriangledown_{0}$ operator will be a morphological rule
without alterations.  The $\bigtriangledown_{1}$ operator will extract
the rules with the alterations in the last letter of the main word, as in
the example above. The $\bigtriangledown$ operator is applied to a pair
of words from the lexicon. First it segments the last $n$ characters of
the shorter word and stores this in the $M$ element of the rule. Then it
tries to segment an affix by subtracting the shorter word without the
mutative ending from the longer word.  If the subtraction results in an
non-empty string it creates a morphological rule by storing the \pos
-class of the shorter word as the $I$-class, the \pos -class of the
longer word as the $R$-class and the segmented affix itself.  For
example:

{\andrei [booked (JJ VBD VBN)]} $\bigtriangledown_{0}$
{\andrei [book (NN VB)]}  \ra  \\
\hspace*{14ex}  $A^{s}:${\andrei [ed (NN VB) (JJ VBD VBN)  ``'']}

{\andrei [advisable (JJ VBD VBN)]} $\bigtriangledown_{1}$
{\andrei [advise (NN VB)]}  \ra  \\
\hspace*{10ex}  $A^{s}:${\andrei [able (NN VB) (JJ VBD VBN)  ``e'']}

The $\bigtriangledown$ operator is applied to all possible lexicon-entry
pairs and if a rule produced by such an application has already been
extracted from another pair, its frequency count ($f$) is incremented.
Thus sets of morphological guessing rules together with their calculated
frequencies are produced.  Next, from these sets of guessing rules we
need to cut out infrequent rules which might bias the further learning
process.  To do that we eliminate all the rules with the frequency $f$
less than a certain threshold $\theta$\footnote{usually we set this
  threshold quite low: 2--4.}.  Such filtering reduces the rule-sets more
than tenfold and does not leave clearly coincidental cases among the
rules.

\subsection{Rule Scoring Phase}
\label{sec:score}

Of course, not all acquired rules are equally good as plausible guesses
about word-classes.  So, for every acquired rule we need to estimate
whether it is an effective rule which is worth retaining in the final
rule-set.  
\ignore{ For such estimation we perform a statistical
  experiment as follows: for every rule we calculate the number of times
  the rule was applied to a word-token from a raw corpus and the number
  of times it gave the right answer. Note that the task of the rule is
  not to disambiguate a word's \pos\ but to provide {\em all} possible
  \pos -tags it can take on. If the rule is correct in the majority of
  times it was applied it is obviously a good rule. If the rule is wrong
  most of the times it is a bad rule which should not be included into
  the final rule-set.  } 
To perform such estimation we take one-by-one
each rule from the rule-sets produced at the rule extraction phase, take
each word-token from the corpus and guess its \pos -set using the rule if
the rule is applicable to the word. For example, if a guessing rule
strips a particular suffix and a current word from the corpus does not
have such suffix we classify these word and rule as incompatible and the
rule as not applicable to that word. If the rule is applicable to the
word we perform lookup in the lexicon and then compare the result of the
guess with the information listed in the lexicon. If the guessed \pos
-set is the same as the \pos -set stated in the lexicon, we count it as
success, otherwise it is failure. Then for each rule we calculate its
score as explained in \cite{Mikheev:1996} using the scoring function as
follows:

$score_{i}= \hat{p}_{i} -
1.65*\sqrt{\frac{\hat{p}_{i}(1-\hat{p}_{i})}{n_{i}}}/(1+\log(|S_{i}|))$

where $\hat{p}$ is the proportion of all positive outcomes ($x$) of the
rule application to the total number of compatible to the rule words
($n$), and $|S|$ is the length of the affix. We also smooth $\hat{p}$ so
as not to have zeros in positive or negative outcome probabilities: 
$\hat{p} = \frac{x+0.5 }{n+1 } $.

Setting the threshold $\theta_{s}$ at a certain level lets only the rules
whose score is higher than the threshold to be included into the final
rule-sets.  The method for setting up the threshold is based on empirical
evaluations of the rule-sets and is described in Section~\ref{sec:eval}.

\subsection{Setting the Threshold}
\label{sec:eval}

The task of assigning a set of \pos -tags to a particular word is
actually quite similar to the task of document categorisation where a
document should be assigned with a set of descriptors which represent its
contents.  The performance of such assignment can be measured in:
 
{\em recall} - the percentage of \pos -tags which the guesser  assigned
correctly  to a word;

{\em precision} - the percentage of \pos -tags the  guesser
assigned correctly  over the total number of \pos -tags
it assigned to the word;

{\em coverage} - the proportion of words which the guesser was able to
classify, but not necessarily correctly.


\ignore{ To see how well a set of guessing rules performs we can compare
  the results of the guessing with the set of \pos -tags which are known
  to be true for the word.  For example, suppose that the word
  ``developed'' is stated in the lexicon as {\andrei (JJ VBD VBN)}.
  Suppose that the guesser categorised it as: {\andrei (JJ NN QL VBD
    VBZ)}.  This guessing will have recall of two {\andrei (JJ VBD)} out
  of three {\andrei (JJ VBD VBN)} or 66\%. 100\% recall would mean that
  the guesser assigned all the correct \pos s but not necessarily only
  the correct. So, for example, if the guesser assigned all existing \pos
  s its recall would have been 100\%. The precision of the guessing in
  the example above will be two {\andrei (JJ VBD)} out of five {\andrei
    (JJ NN QL VBD VBZ)} or 40\%.  100\% precision would mean that the
  guesser didn't assign incorrect \pos s, although not necessarily all
  the correct ones were assigned. So if the guesser assigned only
  {\andrei JJ} its precision would have been 100\%.  
}

There are two types of test-data in use at this stage.  First, we measure
the performance of a guessing rule-set against the actual lexicon: every
word from the lexicon, except for closed-class words and words shorter
than five characters, is guessed by the rule-sets and the results are
compared with the information the word has in the lexicon.  In the second
experiment we measure the performance of the guessing rule-sets against
the training corpus.  For every word we measure its metrics exactly as in
the previous experiment. Then we multiply these measures by the corpus
frequency of this particular word and average them. Thus the most
frequent words have the greatest influence on the final measures.

To extract the best-scoring rule-sets for each acquired set of rules we
produce several final rule-sets setting the threshold $\theta_{s}$ at
different values. For each produced rule-set we record the three metrics
(precision, recall and coverage) and choose the sets with the best
aggregate measures.

\section{Learning Experiment}

\begin{table*}[t]  

\begin{minipage}{\hsize}

\begin{tabular}[t]{||l|lll|lll||}
\hline\hline
Guessing  & \multicolumn{3}{c|}{Lexicon}    &  \multicolumn{3}{c|}{Corpus}     \\
Strategy  & Precision & Recall   & Coverage & Precision & Recall  &  Coverage \\
\hline
Suffix 
(S$_{60}$) & 0.920476 & 0.959087 & 0.373851 & 0.978246 & 0.973537 & 0.29785  \\
Suffix with alt. 
(A$_{80}$) & 0.964433 & 0.97194  & 0.193404 & 0.996292 & 0.991106 & 0.187478 \\
S$_{60}$+
A$_{80}$   & 0.925782 & 0.959568 & 0.4495   & 0.981375 & 0.977098 & 0.370538 \\
A$_{80}$+
S$_{60}$   & 0.928376 & 0.959457 &  0.4495  & 0.981844 & 0.977165 & 0.370538 \\
\hline
Ending 
(E$_{75}$) & 0.666328 & 0.94023  & 0.97741  & 0.755653 & 0.951342 & 0.958852 \\
S$_{60}$+
E$_{75}$   & {\bf 0.728449} 
                      & 0.941157 & 0.9789471& {\bf 0.798186}  
                                                       & 0.947714 & 0.961047 \\
S$_{60}$+A$_{80}$+
E$_{75}$   & 0.739347 & 0.941548 & 0.979181 & 0.805789 & 0.948022 & 0.961047 \\
A$_{80}$+S$_{60}$+
E$_{75}$ & {\bf 0.740538}
                      & 0.941497 & 0.979181 & {\bf 0.805965}
                                                       & 0.948051 & 0.961047 \\
\ignore{
P$_{80}$+A$_{80}$+S$_{60}$+
E$_{75}$ & 0.75103    & 0.942411 & 0.97958  & 0.814592 & 0.958643 & 0.961047 \\
}
\hline\hline
\end{tabular}

\vspace{2ex} 

\caption{ Results of the cascading application of the rule-sets over the
  training lexicon and training corpus. A$_{80}$ - suffixes with
  alterations scored over 80 points, S$_{60}$ - suffixes without
  alterations scored over 60 points, E$_{75}$ - ending-guessing rule-set
  scored over 75 points.  \ignore{ The first part of the table shows that
    the cascading application of the two suffix rule-sets (A$_{80}$ and
    S$_{60}$) increased their joint lexical coverage by about 7-8\%.  The
    second part of the table shows that the cascading application of the
    suffix rules (S$_{60}$) together with the ending-guessing rules
    (E$_{75}$) increased the performance about 5\% in precision and the
    addition of the suffix rules with alterations (A$_{80}$) caused
    further improvement in precision by about 1\%.}}

\longline

\end{minipage}
\end{table*}

One of the most important issues in the induction of guessing rule-sets
is the choice of right data for training.  In our approach, guessing
rules are extracted from the lexicon and the actual corpus frequencies of
word-usage then allow for discrimination between rules which are no
longer productive (but have left their imprint on the basic lexicon) and
rules that are productive in real-life texts.  Thus the major factor in
the learning process is the lexicon - it should be as general as possible
(list {\em all} possible \pos s for a word) and as large as possible,
since guessing rules are meant to capture general language regularities.
The corresponding corpus should include most of the words from the
lexicon and be large enough to obtain reliable estimates of
word-frequency distribution.

We performed a rule-induction experiment using the lexicon and
word-frequencies derived from the Brown Corpus \cite{Francis:1982}.
There are a number of reasons for choosing the Brown Corpus data for
training.  The most important ones are that the Brown Corpus provides a
model of general multi-domain language use, so general language
regularities can be induced from it, and second, many taggers come with
data trained on the Brown Corpus which is useful for comparison and
evaluation. This, however, by no means restricts the described technique
to that or any other tag-set, lexicon or corpus.  Moreover, despite the
fact that the training is performed on a particular lexicon and a
particular corpus, the obtained guessing rules suppose to be domain and
corpus independent and the only training-dependent feature is the tag-set
in use.

Using the technique described above and the lexicon derived from the
Brown Corpus we extracted prefix morphological rules (no alterations),
suffix morphological rules without alterations and ending guessing rules,
exactly as it was done in \cite{Mikheev:1996}.  Then we extracted suffix
morphological rules with alterations in the last letter
($\bigtriangledown_{1}$), which was a new rule-set for the cascading
guesser.  Quite interestingly apart from the expected suffix rules with
alterations as:

[ $S$={\andrei ied} $I$={\andrei (NN, VB)} $R$={\andrei (JJ VBD  VBN)} $M$=y]

which can handle pairs like {\andrei deny \ra denied}, this rule-set was
populated with ``second-order'' rules which describe dependencies between
secondary forms of words. For instance, the rule

[ $S$={\andrei ion} $I$= {\andrei (NNS VBZ)}  $R$={\andrei (NN)}  $M$=s]

says if by deleting the suffix ``ion'' from a word and adding ``s'' to
the end of the result of this deletion we produce a word which is listed
in the lexicon as a plural noun and 3-rd form of a verb {\andrei (NNS
  VBZ)} the unknown word is a noun {\andrei (NN)}. This rule, for
instance, is applicable to word pairs: {\andrei affects \ra affection},
{\andrei asserts \ra assertion}, etc.

Table~1 presents some results of a comparative study of the cascading
application of the new rule-set against the standard rule-sets of the
cascading guesser.  The first part of Table~1 shows the best obtained
scores for the standard suffix rules (S) and suffix rules with
alterations in the last letter (A).  When we applied the two suffix
rule-sets cascadingly their joint lexical coverage increased by about
7-8\% (from 37\% to 45\% on the lexicon and from 30\% to 37\% on the
corpus) while precision and recall remained at the same high level. This
was quite an encouraging result which, actually, agreed with our
prediction.  Then we measured whether suffix rules with alterations (A)
add any improvement if they are used in conjunction with the
ending-guessing rules.  Like in the previous experiment we measured the
precision, recall and coverage both on the lexicon and on the corpus.
The second part of Table~1 shows that simple concatenative suffix rules
(S$_{60}$) improved the precision of the guessing when they were applied
before the ending-guessing rules (E$_{75}$) by about 5\%. Then we
cascadingly applied the suffix rules with alterations (A$_{80}$) which
caused further improvement in precision by about 1\%.

\begin{table*}[t]  
\begin{minipage}{\hsize}
\begin{tabular}[t]{||l|l||l|l|l|l||l|l||}
\hline\hline
Lexicon & Guessing       & Total & Unkn. & Total   & Unkn.  &  Total  & Unkn. \\
        & strategy       & words & words & mistag. & mistag.&  Score  & Score \\
\hline\hline
Full  & standard: P+S+E  & 5,970 & 347   & 292     & 33     & 95.1\%  & 90.5\% \\
Full  & with new: P+A+S+E& 5,970 & 347   & 292     & 33     & 95.1\%  & 90.5\% \\
\hline
Small & standard: P+S+E  & 5,970 & 2,215 & 332     & 309    & 94.44\% & 86.05\% \\
Small & with new: P+A+S+E& 5,970 & 2,215 & 311     & 288    & 94.79\% & 87.00\% \\
\hline\hline
\end{tabular}

\vspace{2ex} 

\label{tab:tag-2}

\caption{ Results of tagging  a text using the standard
  Prefix+Suffix+Ending cascading guesser and the guesser with the
  additional rule-set of suffixes-with-Alterations. For each of these
  cascading guessers two tagging experiments were performed: the tagger
  was equipped with the full Brown Corpus lexicon and with the small
  lexicon of closed-class and short words (5,465 entries).
  \ignore{Whereas the additional rule-set did not cause any noticeable
    improvement when tagging with the full-fledged lexicon, we measured
    about 1\% improvement in tagging accuracy on unknown words when
    tagging with the small lexicon.} }

\longline
\end{minipage}
\end{table*}

After obtaining the optimal rule-sets we performed the same experiments
on a word-sample which was not included into the training lexicon and
corpus.  We gathered about three thousand words from the lexicon
developed for the Wall Street Journal corpus\footnote{these words were
  not listed in the training lexicon} and collected frequencies of these
words in this corpus.  At this test-sample evaluation we obtained similar
metrics apart from the coverage which dropped by about 7\% for both kinds
of suffix rules.  This, actually, did not come as a surprise, since many
main forms required by the suffix rules were missing in the lexicon.

\section{Evaluation}

The direct performance measures of the rule-sets gave us the grounds for
the comparison and selection of the best performing guessing rule-sets.
The task of unknown word guessing is, however, a subtask of the overall
part-of-speech tagging process. Thus we are mostly interested in how the
advantage of one rule-set over another will affect the tagging
performance.  So, we performed an independent evaluation of the impact of
the word guessing sets on tagging accuracy. In this evaluation we used
the cascading application of prefix rules, suffix rules and
ending-guessing rules as described in \cite{Mikheev:1996}. We measured
whether the addition of the suffix rules with alterations increases the
accuracy of tagging in comparison with the standard rule-sets.  In this
experiment we used a tagger which was a {\sc c}++ re-implementation of
the {\sc lisp} implemented HMM Xerox tagger described in
\cite{Kupiec:1992} trained on the Brown Corpus.  For words which failed
to be guessed by the guessing rules we applied the standard method of
classifying them as common nouns (NN) if they are not capitalised inside
a sentence and proper nouns (NP) otherwise.
 
In the evaluation of tagging accuracy on unknown words we payed attention
to two metrics.  First we measure the accuracy of tagging solely on
unknown words:

$Unkown Score = \frac{Correctly Tagged Unkown Words}{Total Unknown Words}$

This metric gives us the exact measure of how the tagger has done when
equipped with different guessing rule-sets.  In this case, however, we do
not account for the known words which were mis-tagged because of the
unknown ones. To put a perspective on that aspect we measure the overall
tagging performance:

$TotalScore = \frac{Correctly Tagged Words}{Total Words}$

To perform such evaluation we tagged several texts of different origins,
except ones from the Brown Corpus. These texts were not seen at the
training phase which means that neither the tagger nor the guesser had
been trained on these texts and they naturally had words unknown to the
lexicon. For each text we performed two tagging experiments. In the first
experiment we tagged the text with the full-fledged Brown Corpus lexicon
and hence had only those unknown words which naturally occur in this
text. In the second experiment we tagged the same text with the lexicon
which contained only closed-class\footnote{articles, prepositions,
  conjunctions, etc.} and short\footnote{shorter than 5 characters}
words.  This small lexicon contained only 5,456 entries out of 53,015
entries of the original Brown Corpus lexicon.  All other words were
considered as unknown and had to be guessed by the guesser. In both
experiments we measured tagging accuracy when tagging with the guesser
equipped with the standard Prefix+Suffix+Ending rule-sets and with the
additional rule-set of suffixes with alterations in the last letter.

Table~2 presents some results of a typical example of such experiments.
There we tagged a text of 5,970 words.  This text was detected to have
347 unknown to the Brown Corpus lexicon words and as it can be seen the
additional rule-set did not cause any improvement to the tagging
accuracy.  Then we tagged the same text using the small lexicon.  Out of
5,970 words of the text, 2,215 were unknown to the small lexicon. Here we
noticed that the additional rule-set improved the tagging accuracy on
unknown words for about 1\%: there were 21 more word-tokens tagged
correctly because of the additional rule-set.  Among these words were:
``classified'', ``applied'', ``tries'', ``tried'', ``merging'',
``subjective'', etc.

\section{Discussion and Conclusion}

The target of the research reported in this paper was to incorporate the
learning of morphological word-\pos\ guessing rules which do not obey
simple concatenations of main words with affixes into the learning
paradigm proposed in \cite{Mikheev:1996}. To do that we extended the data
structures and the algorithms for the guessing-rule application to handle
the mutations in the last $n$ letters of the main words.  Thus simple
concatenative rules naturally became a subset of the mutative rules --
they can be seen as mutative rules with the zero mutation, i.e. when the
$M$ element of the rule is empty.  Simple concatenative rules, however,
are not necessarily regular morphological rules and quite often they
capture other non-linear morphological dependencies.  For instance,
consonant doubling is naturally captured by the affixes themselves and
obey simple concatenations, as, for example, describes the suffix rule
$A^{s}$:

[ $S$={\andrei ging} $I$={\andrei (NN VB)} $R$={\andrei (JJ NN VBG)} $M$=``''] 

rule, for example, will work for word pairs like {\andrei tag - tagging}
or {\andrei dig - digging}.  Note that here we don't specify the
prerequisites for the stem-word to have one syllable and end with the
same consonant as in the beginning of the affix. Our task here is not to
provide a precise morphological description of English but rather to
support computationally effective \pos -guessings, by employing some
morphological information. So, instead of using a proper morphological
processor, we adopted an engineering approach which is argued for in
\cite{Mikheev:1995}. There is, of course, nothing wrong with
morphological processors perse, but it is hardly feasible to re-train
them fully automatically for new tag-sets or to induce new rules.  Our
shallow technique on the contrary allows to induce such rules completely
automatically and ensure that these rules will have enough discriminative
features for robust guessings.  In fact, we abandoned the notion of
morpheme and are dealing with {\em word segments} regardless of whether
they are ``proper'' morphemes or not.  So, for example, in the rule above
``ging'' is considered as a suffix which in principle is not right: the
suffix is ``ing'' and ``g'' is the dubbed consonant. Clearly, such
nuances are impossible to learn automatically without specially prepared
training data, which is denied by the technique in use.  On the other
hand it is not clear that this fine-grained information will contribute
to the task of morphological guessing. The simplicity of the proposed
shallow morphology, however, ensures fully automatic acquisition of such
rules and the empirical evaluation presented in section~\ref{sec:eval}
confirmed that they are just right for the task: precision and recall of
such rules were measured in the range of 96-99\%.

The other aim of the research reported here was to assess whether
non-concatenative morphological rules will improve the overall
performance of the cascading guesser. As it was measured in
\cite{Mikheev:1996} simple concatenative prefix and suffix morphological
rules improved the overall precision of the cascading guesser by about
5\%, which resulted in 2\% higher accuracy of tagging on unknown words.
The additional rule-set of suffix rules with one letter mutation caused
some further improvement. The precision of the guessing increased by
about 1\% and the tagging accuracy on a very large set of unknown words
increased by about 1\%.  In conclusion we can say that although the
ending-guessing rules, which are much simpler than morphological rules,
can handle words with affixes longer than two characters almost equally
well, in the framework of \pos -tagging even a fraction of percent is an
important improvement.  Therefore the contribution of the morphological
rules is valuable and necessary for the robust \pos -tagging of
real-world texts.
 
\ignore{ So, in conclusion we can say that if we knew that the
  performance increase would be as low as that we wouldn't probably have
  ventured to extract the rules with the mutations, but since their
  learning has been naturally incorporated into the overall learning
  process and they cause some improvement we decided to leave them in the
  cascading guesser following the motto ``every little helps''.  }

\end{document}